%% file: ms.tex
\documentclass[journal]{IEEEtran}

\usepackage{cite}
\usepackage{array}
\usepackage{graphicx}
\usepackage{url}
\usepackage{amsfonts}
\usepackage{multirow}
\usepackage{subcaption}
\usepackage{tabularx}
\usepackage{amsmath}
\usepackage{comment}
\usepackage{array}
\usepackage{amssymb}
\usepackage{import}
\usepackage[leftcaption]{sidecap}
\usepackage{algorithmic}
\usepackage{hyperref}

\DeclareRobustCommand*{\IEEEauthorrefmark}[1]{%
  \raisebox{0pt}[0pt][0pt]{\textsuperscript{\footnotesize #1}}%
}

\begin{document}
\title{2018 Robotic Scene Segmentation Challenge}
\author{
    \IEEEauthorblockN{
      M. ~Allan\IEEEauthorrefmark{1}, 
      S. ~Kondo\IEEEauthorrefmark{2}, 
      S. ~Bodenstedt\IEEEauthorrefmark{3}, 
      S. ~Leger\IEEEauthorrefmark{3},
      R. ~Kadkhodamohammadi\IEEEauthorrefmark{4},
      I. ~Luengo\IEEEauthorrefmark{4},
      F. ~Fuentes\IEEEauthorrefmark{4},
      E. ~Flouty\IEEEauthorrefmark{4},
      A. ~Mohammed\IEEEauthorrefmark{5},
      M. ~Pedersen\IEEEauthorrefmark{5},
      A. ~Kori\IEEEauthorrefmark{6},
      V. ~Alex\IEEEauthorrefmark{6},
      G. ~Krishnamurthi\IEEEauthorrefmark{6},
      D. ~Rauber\IEEEauthorrefmark{7},
      R. ~Mendel\IEEEauthorrefmark{7},
      C. ~Palm\IEEEauthorrefmark{7},
      S. ~Bano\IEEEauthorrefmark{8},
      G. ~Saibro\IEEEauthorrefmark{9},
      C. S. ~Shih\IEEEauthorrefmark{10},
      H. A. ~Chiang\IEEEauthorrefmark{10},
      J. ~Zhuang\IEEEauthorrefmark{11},
      J. ~Yang\IEEEauthorrefmark{11},
      V. ~Iglovikov\IEEEauthorrefmark{12},
      A. ~Dobrenkii\IEEEauthorrefmark{12},
      X. ~Liu\IEEEauthorrefmark{13},
      C. ~Gao\IEEEauthorrefmark{13},
      M. ~Unberath\IEEEauthorrefmark{13},
      M. ~Reddiboina\IEEEauthorrefmark{14},
      A. ~Reddy\IEEEauthorrefmark{14},
      M. ~Kim\IEEEauthorrefmark{16},
      C. ~Kim\IEEEauthorrefmark{16}, 
      C. ~Kim\IEEEauthorrefmark{16}, 
      H. ~Kim\IEEEauthorrefmark{16}, 
      G. ~Lee\IEEEauthorrefmark{16}, 
      I. Ullah\IEEEauthorrefmark{16},
      M. Luna\IEEEauthorrefmark{16},
      S. H. Park\IEEEauthorrefmark{16},
      M. ~Azizian\IEEEauthorrefmark{1},
      D. ~Stoyanov\IEEEauthorrefmark{8},\IEEEauthorrefmark{4}
      L. ~Maier-Hein\IEEEauthorrefmark{15},
      S. ~Speidel\IEEEauthorrefmark{3}\\
    }
    \IEEEauthorblockA{
      \IEEEauthorrefmark{1}Intuitive Inc.,
      \IEEEauthorrefmark{2}Konika Minola Inc.,
      \IEEEauthorrefmark{3}National Center for Tumor Diseases (NCT),
      \IEEEauthorrefmark{4}Digital Surgery Ltd.,
      \IEEEauthorrefmark{5}Norwegian University of Science and Technology, 
      \IEEEauthorrefmark{6}Indian Institute of Technology Madras, 
      \IEEEauthorrefmark{7}OTH Regensburg, 
      \IEEEauthorrefmark{8}Wellcome/EPSRC Centre for Interventional and Surgical Sciences (WEISS) UCL,
      \IEEEauthorrefmark{9}IRCAD, 
      \IEEEauthorrefmark{10}National Taiwan University,
      \IEEEauthorrefmark{11}Yale University,
      \IEEEauthorrefmark{12}ODS.AI,
      \IEEEauthorrefmark{13}Johns Hopkins University,
      \IEEEauthorrefmark{14}Rediminds Inc.,
      \IEEEauthorrefmark{15}German Cancer Research Center (DKFZ),
      \IEEEauthorrefmark{16}Daegu Gyeongbuk Institute of Science and Technology
  }
}
\maketitle

\begin{abstract}

In 2015 we began a sub-challenge at the EndoVis workshop at MICCAI in Munich using endoscope images of ex-vivo tissue with automatically generated annotations from robot forward kinematics and instrument CAD models. However, the limited background variation and simple motion rendered the dataset uninformative in learning about which techniques would be suitable for segmentation in real surgery. In 2017, at the same workshop in Quebec we introduced the robotic instrument segmentation dataset with 10 teams participating in the challenge to perform binary, articulating parts and type segmentation of da Vinci$^{\tiny \textregistered}$ instruments. This challenge included realistic instrument motion and more complex porcine tissue as background and was widely addressed with modifications on U-Nets and other popular CNN architectures \cite{allan_endovis_2017}.

In 2018 we added to the complexity by introducing a set of anatomical objects and medical devices to the segmented classes. To avoid over-complicating the challenge, we continued with porcine data which is dramatically simpler than human tissue due to the lack of fatty tissue occluding many organs. 

\end{abstract}

\input{tex/introduction}
\input{tex/data}
\input{tex/methods}

\input{tex/results}

\bibliographystyle{IEEEtran}
\bibliography{lib}

\end{document}

%% file: tex/introduction.tex
\section{Introduction}

Robot-assisted minimally invasive surgery (MIS) has revolutionized patient care, bringing the advantages of laparoscopic surgery such as trauma reduction and shorter recovery times to an increased number of procedures and patients. This has been achieved in part by dramatically improving the surgeon's precision and control over the anatomy with dexterous articulated instruments and high fidelity 3D vision \cite{palep_robot_2009}. 

\begin{figure}[t]
\centering
\includegraphics[width=\columnwidth]{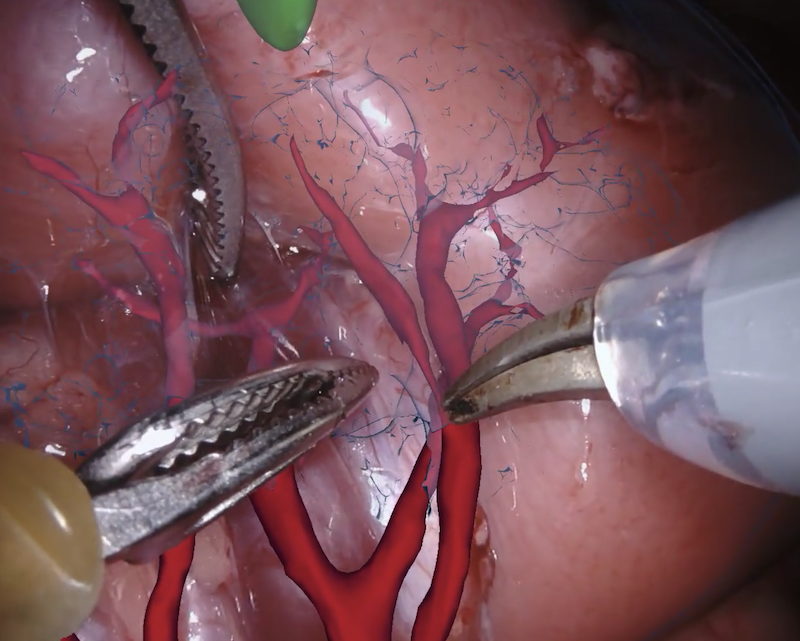}
\caption{\label{fig:overlay}An example of the fusion of segmented blood vessels from pre-operative CT imaging with the endoscopic view.}
\end{figure}

The next paradigm in improving surgeon capabilities is to extend their perception through the fusion of multiple data sources, such as pre- and intra-operative medical imaging modalities, with the endoscopic view (see Figure \ref{fig:overlay}). To display this type of data selectively and intelligently, avoiding cluttering the surgeon's view with information that is not valuable, a critical step is to understand what object are currently in view of the endoscope and which parts of the image they represent. This can lead to a higher-level understanding of what type of anatomy a surgeon is currently interacting with or observing or alternatively which task they are performing. 

To achieve this, a pixel-wise segmentation of the images captured by the endoscopic camera is required and the state-of-the-art for this type of technique is to use deep convolutional neural networks (CNNs) \cite{chen2018encoderdecoder, resnext_xie_2017}. These models require huge amounts of data to train and evaluate effectively and a significant limitation within the medical community is lack of high quality labelled data. This has led to the performance advances that have been demonstrated across the mainstream computer vision community to not be frequently observed in medicine, particularly surgery. 

To address this shortcoming, in 2017 we released a dataset and challenge to assess the state-of-the-art for surgical image segmentation \cite{allan_endovis_2017}. 10 datasets of porcine endoscope images where the da Vinci$^{\tiny \textregistered}$ instruments were segmented into different articulating parts as well as providing labels for the different instrument types. 10 different teams submitted methods and the challenge was won by a team using a modified U-Net architecture \cite{shvets_automatic_2018}. The accuracy of the binary and parts based methods for many of the participants exceeded 0.7 mIoU. Although this score would not be sufficient to consider the problem solved, the results on instruments demonstrated that the scope could be expanded to include more classes without overcomplicating the problem. This led to the creation of the robotic instrument segmentation sub-challenge of the Endoscopic Vision (EndoVis) challenge\footnote{\url{https://endovissub2018-roboticscenesegmentation.grand-challenge.org}} at MICCAI 2018.

%% file: tex/data.tex
\section{Data}

\subsection{Challenge Overview}

The goal for participants was to perform semantic segmentation of surgical images into a set of medical device classes and a set of anatomical classes. The medical devices were separated into da Vinci instruments, using the same shaft, wrist and jaws division that were used in the 2017 challenge; drop-in ultrasound probes; suturing needles; suturing thread; suction-irrigation devices and surgical clips. This set comprised all non-biological objects that appeared in the images. The anatomical classes were the kidney parenchyma; the kidney fascia and perinephric fat, which we termed `covered kidney', and small intestine. All other anatomical objects in the scene were grouped into a background class. 

\begin{figure}[h!]
\centering
\begin{subfigure}[b]{0.48\columnwidth}
\includegraphics[width=\textwidth]{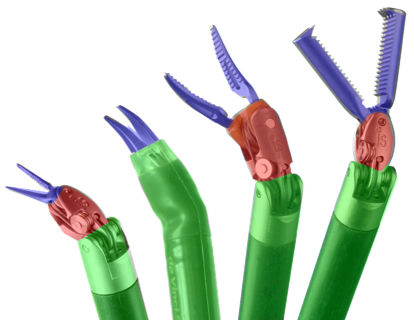}
\caption{\label{fig:instruments}}
\end{subfigure}
\hfill
\begin{subfigure}[b]{0.48\columnwidth}
\includegraphics[width=\textwidth]{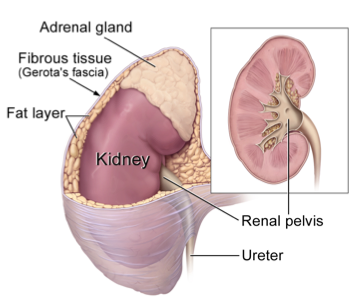}
\caption{\label{fig:kidney}}
\end{subfigure}
\hfill
\\
\begin{subfigure}[b]{0.23\columnwidth}
\includegraphics[width=\textwidth]{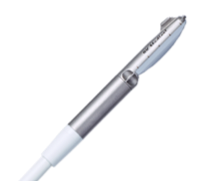}
\caption{\label{fig:us_probe}}
\end{subfigure}
\hfill
\begin{subfigure}[b]{0.23\columnwidth}
\includegraphics[width=\textwidth]{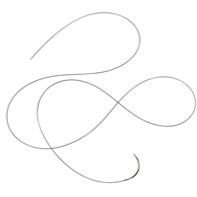}
\caption{\label{fig:needle_thread}}
\end{subfigure}
\hfill
\begin{subfigure}[b]{0.23\columnwidth}
\includegraphics[width=\textwidth]{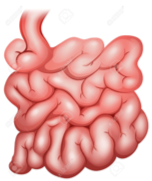}
\caption{\label{fig:small_intestine}}
\end{subfigure}
\hfill
\begin{subfigure}[b]{0.23\columnwidth}
\includegraphics[width=\textwidth]{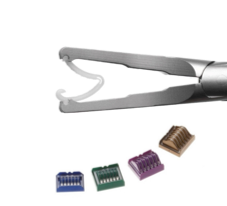}
\caption{\label{fig:clips}}
\end{subfigure}
\caption{\label{fig:classes}The different classes to be segmented in our challenge. The instrument and ultrasound probes were common classes with our 2017 dataset. All other classes were new to this challenge.}
\end{figure}

\subsection{Data Collection}

The entire challenge dataset was made up of 19 sequences which were divided into 15 training sets and 4 test sets. Each sequence came from a single porcine training procedure recorded on da Vinci X or Xi system using specialized recording hardware. Sections of the procedure which contained significant camera motion or tissue interaction were extracted and subsampled to 1 Hz. Similar frames were manually removed until the sequence contained 300 frames. Each frame consists of a stereo pair with SXGA resolution $1280\times1024$ and intrinsic and extrinsic camera calibrations that were acquired during endoscope manufacture.

\begin{figure*}[tb]
\centering
\begin{subfigure}[b]{0.50\columnwidth}
\includegraphics[width=\textwidth]{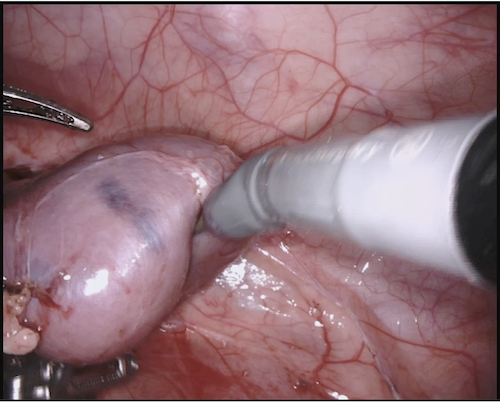}
\end{subfigure}
\hfill
\begin{subfigure}[b]{0.50\columnwidth}
\includegraphics[width=\textwidth]{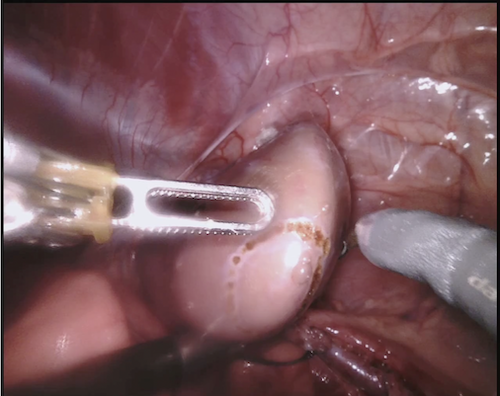}
\end{subfigure}
\hfill
\begin{subfigure}[b]{0.50\columnwidth}
\includegraphics[width=\textwidth]{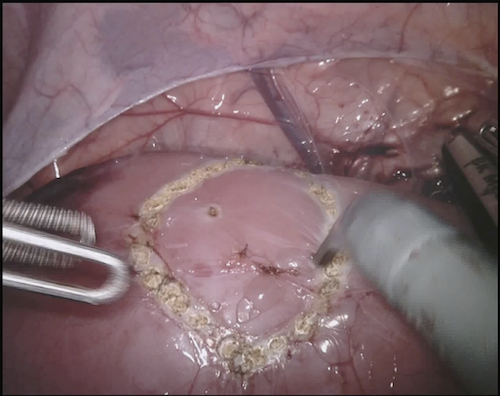}
\end{subfigure}
\hfill
\begin{subfigure}[b]{0.50\columnwidth}
\includegraphics[width=\textwidth]{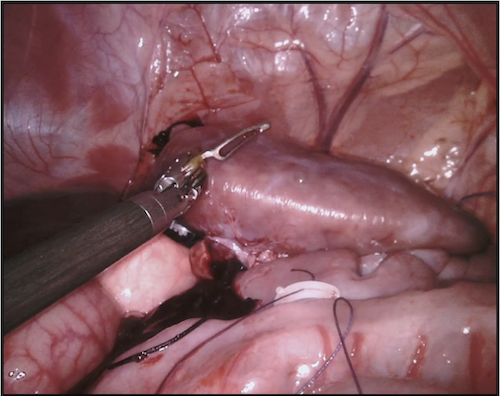}
\end{subfigure}
\hfill
\\
\begin{subfigure}[b]{0.50\columnwidth}
\includegraphics[width=\textwidth]{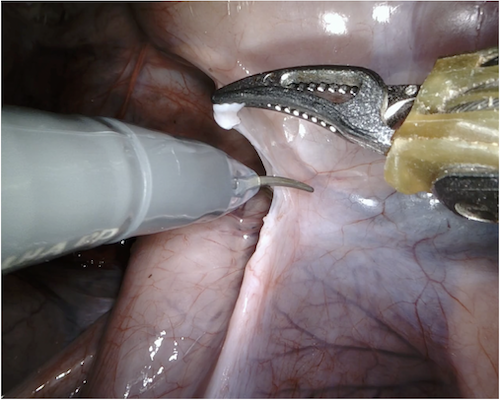}
\end{subfigure}
\hfill
\begin{subfigure}[b]{0.50\columnwidth}
\includegraphics[width=\textwidth]{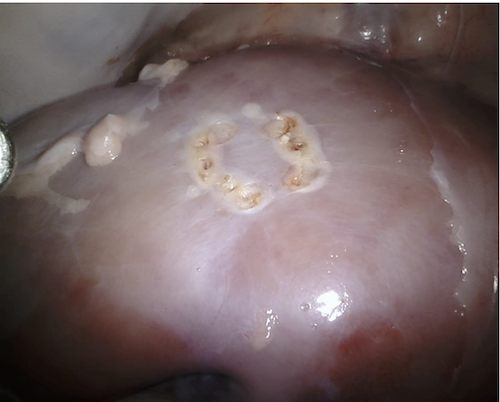}
\end{subfigure}
\hfill
\begin{subfigure}[b]{0.50\columnwidth}
\includegraphics[width=\textwidth]{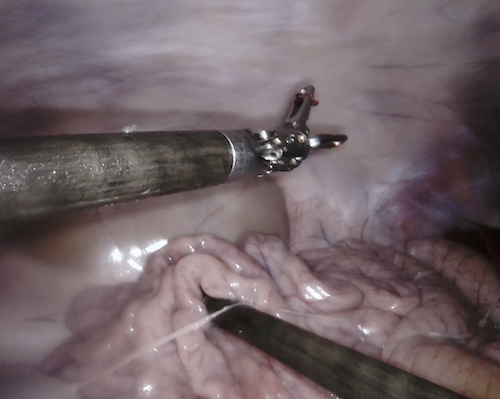}
\end{subfigure}
\hfill
\begin{subfigure}[b]{0.50\columnwidth}
\includegraphics[width=\textwidth]{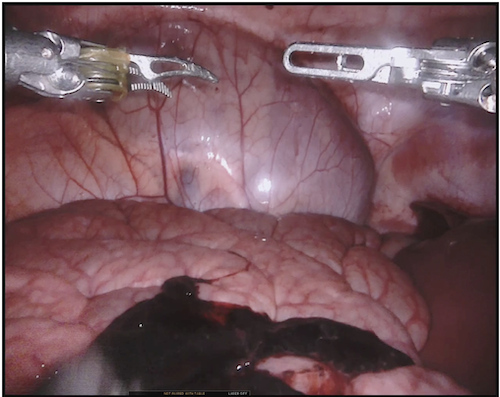}
\end{subfigure}
\caption{\label{fig:training_data} Example frames from the training datasets in order from left to right: Dataset 1, 4, 5, 7, 9, 12, 13, 14.}
\end{figure*}

\subsection{Data Annotation}

The data was annotated by a team of trained technicians at Intuitive in Sunnyvale, CA using in-house software to generate polygons around each semantic class. Quality control was provided by in-house veterinarians. We labelled only the left eye in the stereo pair to reduce annotation time. 

Annotating anatomical data introduced multiple new challenges compared with annotating instruments in our previous challenges where the object classes are very clearly defined. Anatomical data requires a much more complex labelling protocol to resolve ambiguities and achieve a consistent labelling while still respecting higher level objectives for building the segmentation system. 

\begin{figure}[tb]
\centering
\begin{subfigure}[b]{0.48\columnwidth}
\includegraphics[width=\textwidth]{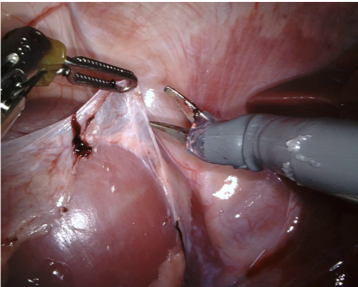}
\end{subfigure}
\hfill
\begin{subfigure}[b]{0.48\columnwidth}
\includegraphics[width=\textwidth]{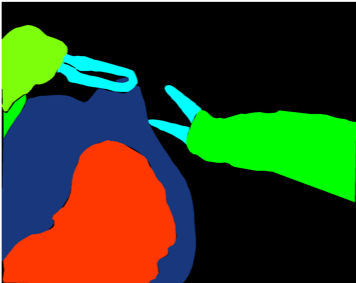}
\end{subfigure}
\caption{\label{fig:ambiguous_label}As the fascia is stretched off the kidney, the case of exactly where the border of the covered kidney label should now lie is fairly complex to describe in a consistent way.}
\end{figure}

For instance, an anatomical segmentation is likely useful if it can identify gross structures such as organs, yet in surgery these structures are often partially covered in connective tissue and fat. To address this issue, we introduced the label `covered kidney' as we hoped to provide special treatment to connective tissue and fat that lies on top of an important anatomical structure by effectively combining two labels together to create a new label. However, this type of label can create complications to describe within a consistent protocol. Figure \ref{fig:ambiguous_label} shows a situation where the fascia is stretched so that it temporarily no longer lies on the tissue surface. An additional complication is that for many camera views of anatomical structures, it may be difficult or impossible for a skilled annotator to make identifications. Instead they need to see an extended sequence of images where the structure can be viewed from different angles and distances. This is complicated by the fact that many out-of-the-box annotation tools consider images independently and do not provide any way to easily view a video sequence within the annotation workflow. 

\begin{figure*}[tb]
\centering
\begin{subfigure}[b]{0.50\columnwidth}
\includegraphics[width=\textwidth]{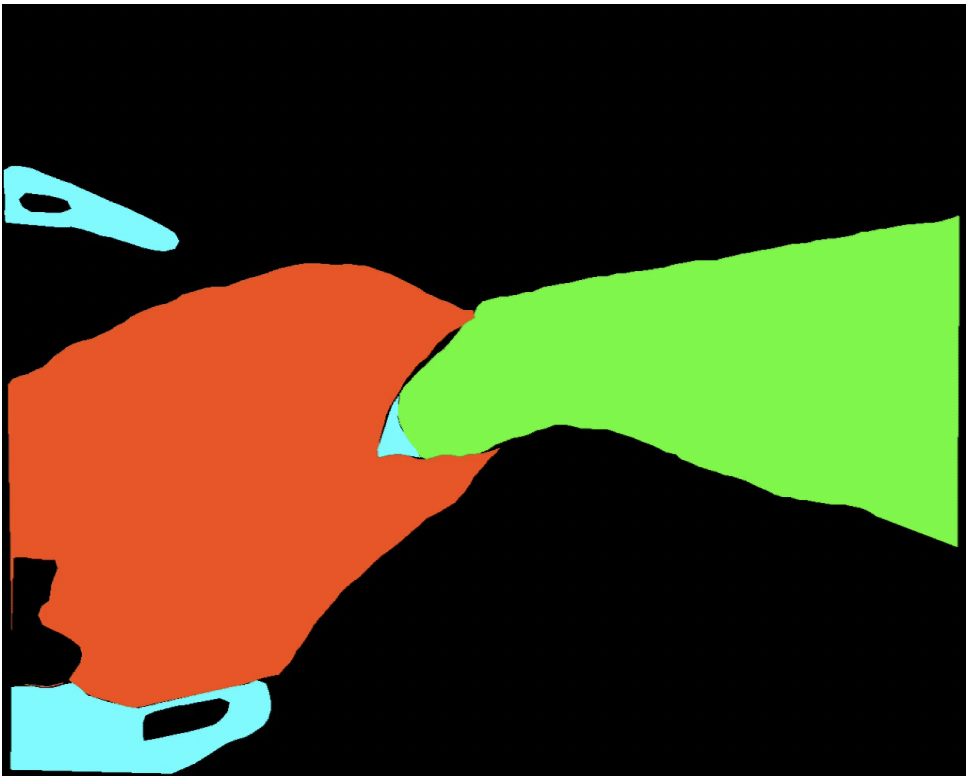}
\end{subfigure}
\hfill
\begin{subfigure}[b]{0.50\columnwidth}
\includegraphics[width=\textwidth]{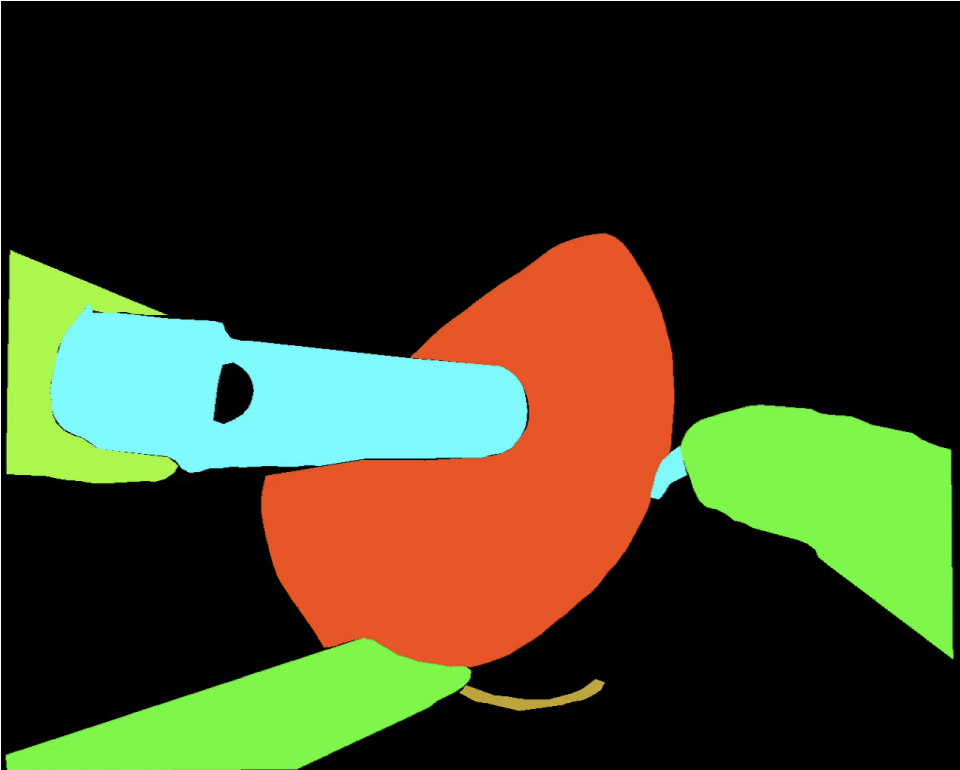}
\end{subfigure}
\hfill
\begin{subfigure}[b]{0.50\columnwidth}
\includegraphics[width=\textwidth]{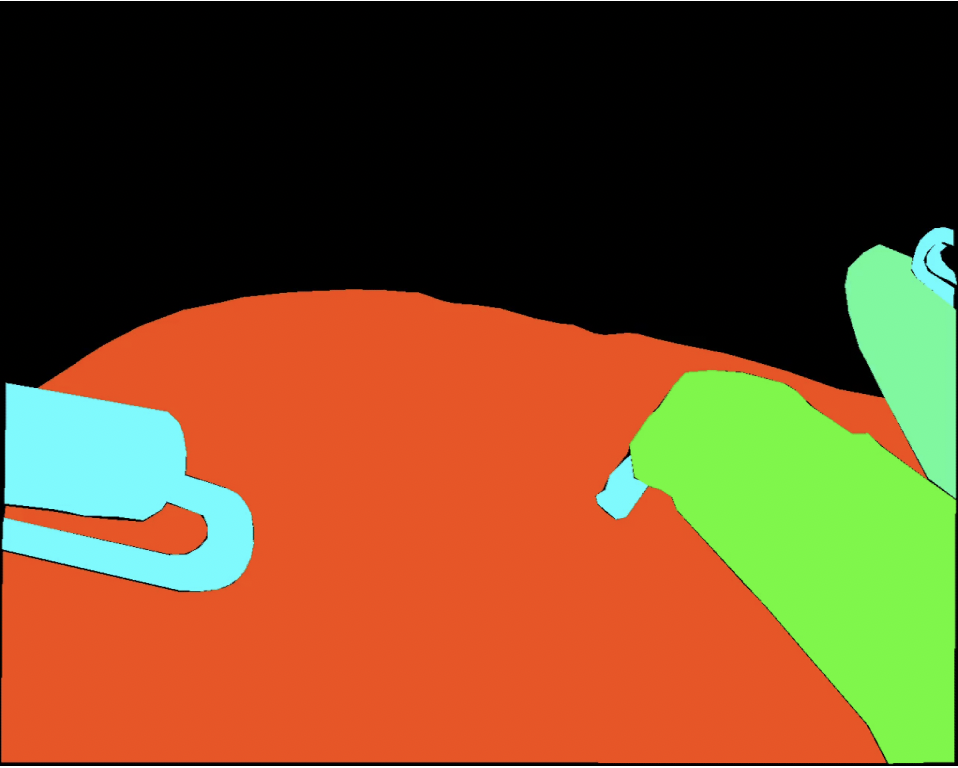}
\end{subfigure}
\hfill
\begin{subfigure}[b]{0.50\columnwidth}
\includegraphics[width=\textwidth]{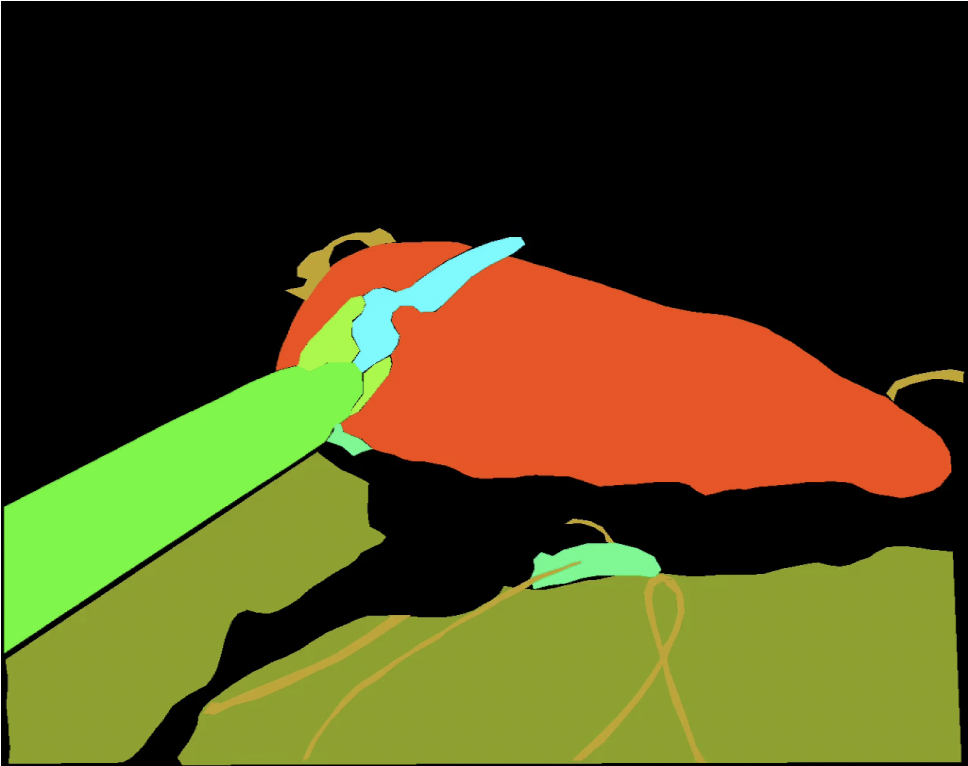}
\end{subfigure}
\hfill
\\
\begin{subfigure}[b]{0.50\columnwidth}
\includegraphics[width=\textwidth]{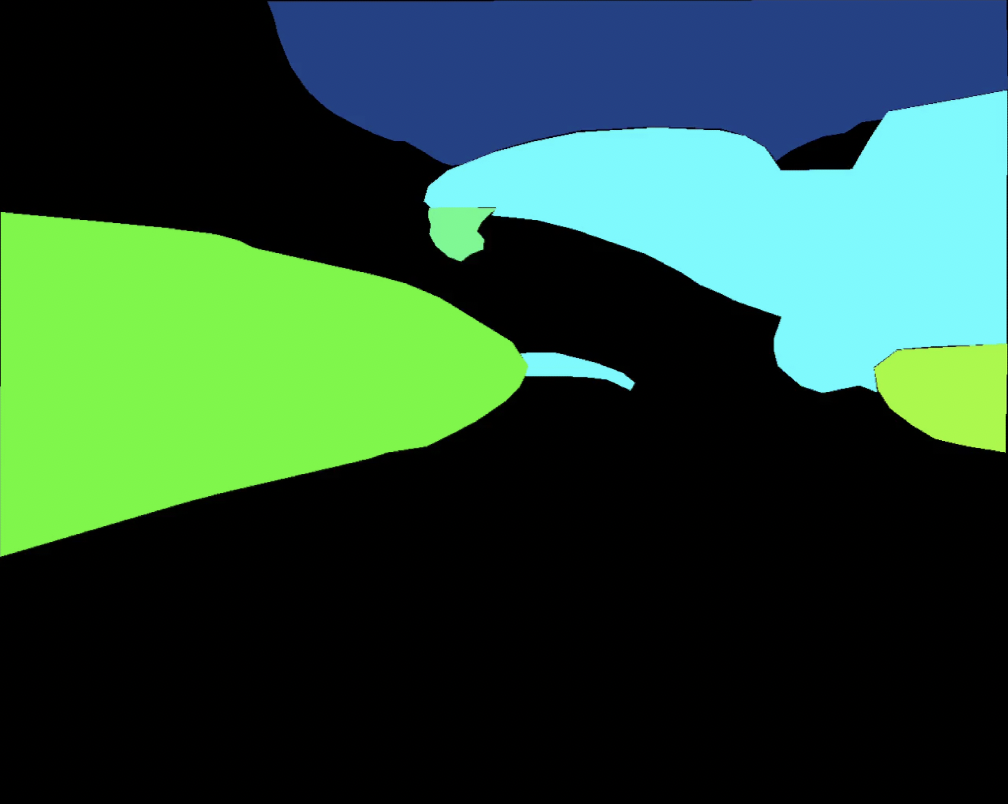}
\end{subfigure}
\hfill
\begin{subfigure}[b]{0.50\columnwidth}
\includegraphics[width=\textwidth]{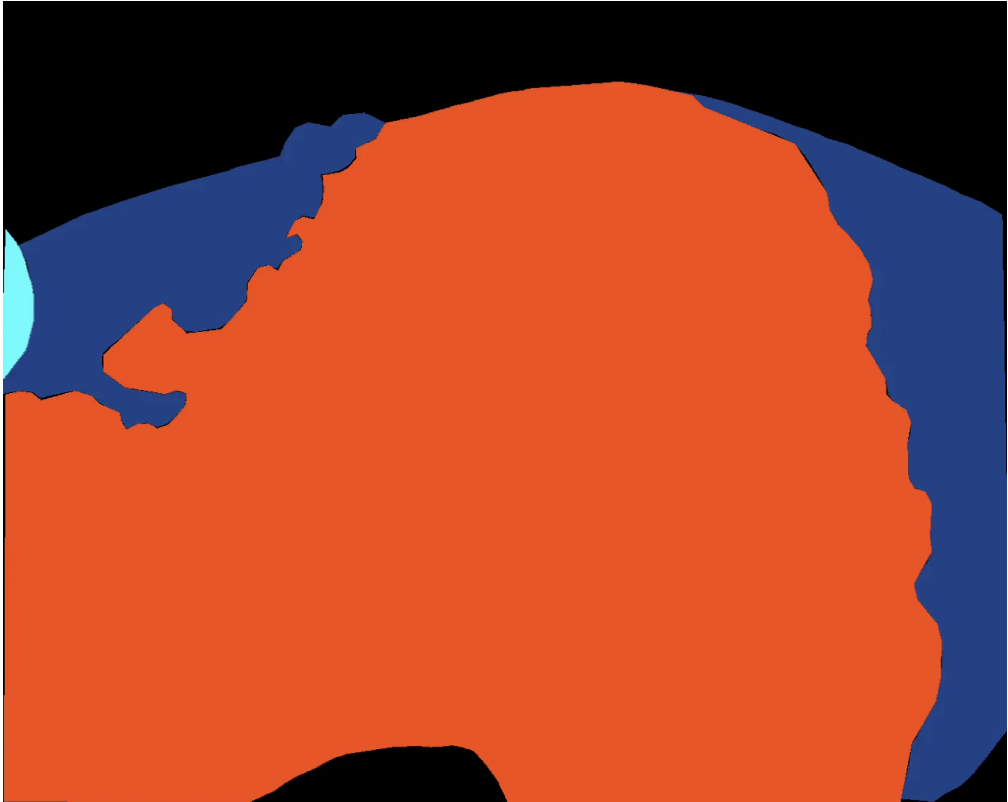}
\end{subfigure}
\hfill
\begin{subfigure}[b]{0.50\columnwidth}
\includegraphics[width=\textwidth]{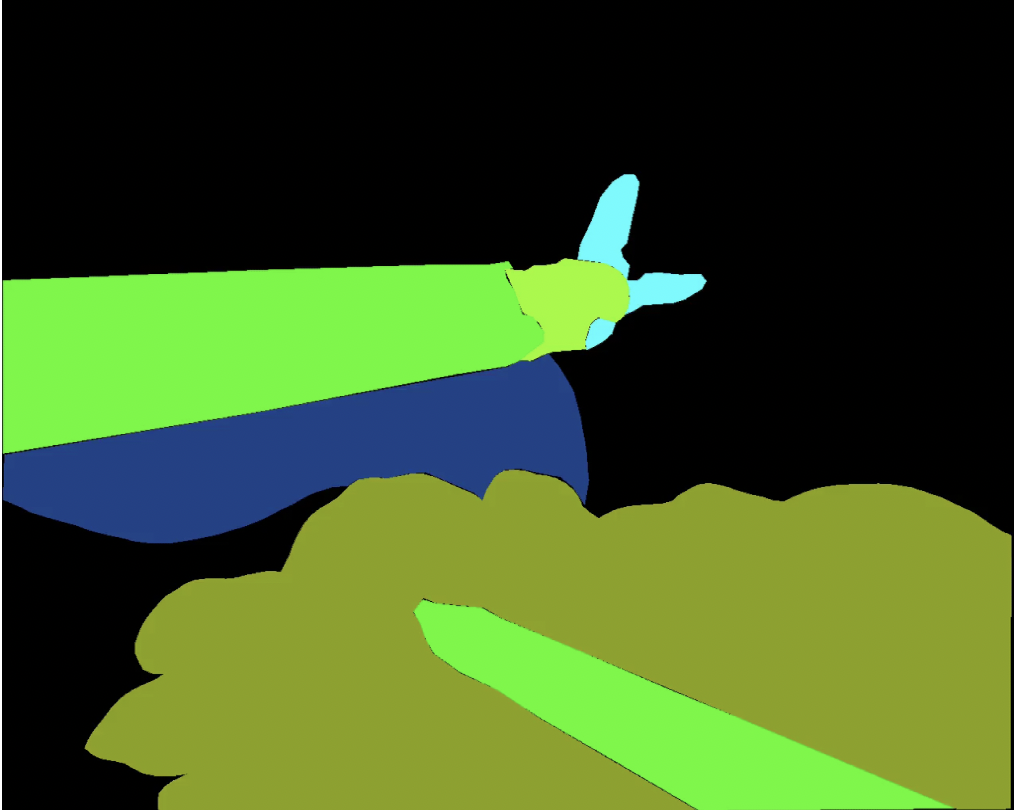}
\end{subfigure}
\hfill
\begin{subfigure}[b]{0.50\columnwidth}
\includegraphics[width=\textwidth]{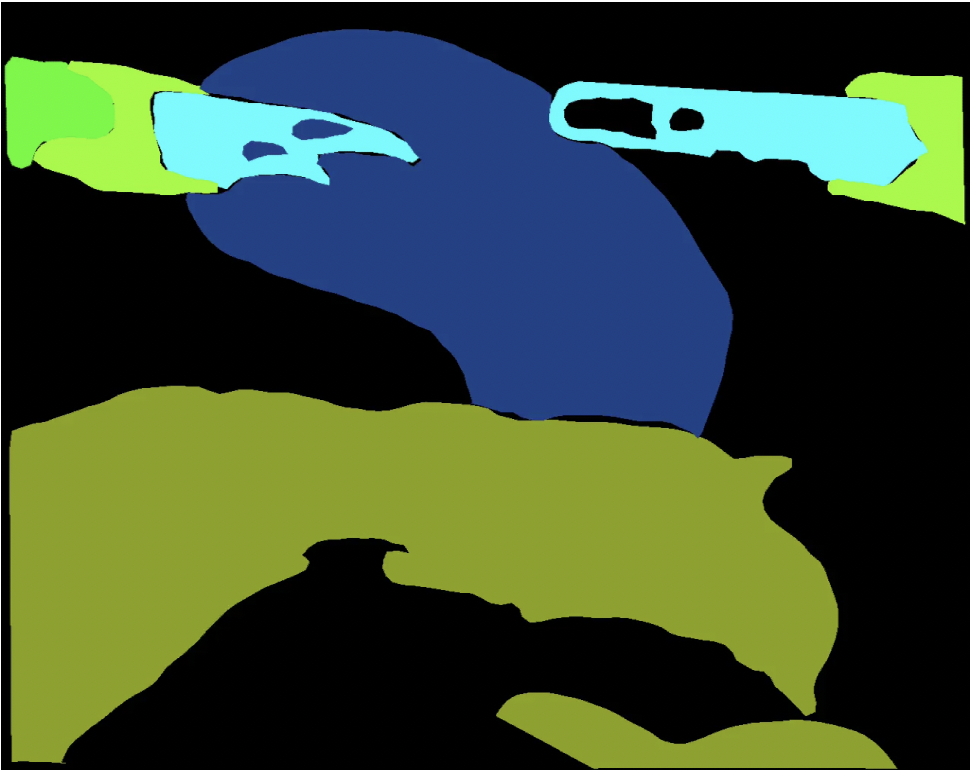}
\end{subfigure}
\caption{\label{fig:training_data} Example annotations from the training datasets in order from left to right: Dataset 1, 4, 5, 7, 9, 12, 13, 14.}
\end{figure*}

%% file: tex/methods.tex
\section{Participating Methods}

\subsection{Konika Minola}

Method 1 was from Satoshi Kondo of Konika Minola Inc, Japan. They used a ResNeXt-101 \cite{resnext_xie_2017} architecture with Squeeze-Excitation blocks \cite{hu2017squeezeandexcitation} and pre-training on the ImageNet dataset\cite{imagenet_deng_2009}. The images are downsampled to half resolution for training. 200 epochs are used with stochastic gradient descent, weighted cross entropy loss and a learning rate of 0.1 using cosine annealing. Translation, rotation, resizing, flips and contrast are the augmentation set and no post-processing is applied to the image. 

\subsection{National Center for Tumor Diseases}

Method 2 was from Sebastian Bodenstedt, Sefan Leger and Stephanie Speidel at the National Center for Tumor Diseases, Dresden, Germany. Their method was to use a U-Net \cite{ronneberger_unet_2015} style architecture with a VGG 19 encoder \cite{simonyan2014deep} pre-trained on ImageNet. Data augmentation was applied through hue, saturation, brightness and contrast jitter. 

\subsection{Digital Surgery}

Method 3 was from Rahim Kadkhodamohammadi, Imanol Luengo, Felix Fuentes
Evangello Flouty and Danail Stoyanov at Digital Surgery Ltd., UK. Their model was based on DeepLab V3+ \cite{chen2018encoderdecoder}. This uses multi-scale feature extraction using Xception \cite{chollet2016xception} and atrous convolution followed by deconvolution layers to predict class labels. They predicted two sets of scoremaps: the 10 classes provided with the dataset and 5 classes constructed by merging semantically related classes. Their network was trained by optimizing the loss:
\begin{equation}
\begin{split}
\mathcal{L} =  \sum_{i=0}^{N}\big[ y_{i}log(\hat{y_{i}}) + (1-y_{i})\log(1-\hat{y_{i}}) \big] \\
\times \big[(d_{max} - \min_{b_{j} \in B} d(i, b_{j}))/2d_{max} + 0.5\big]
\end{split}
\end{equation}
where $B$ is the set of boundary pixels, $N$ is the number of pixels in the image, $d(a, b)$ defines the Euclidean distance and $d_{max}$ is the max distance to the boundaries in the batch. Their inference time is 1 FPS on a NVIDIA GeForce 1080 Ti.

\subsection{Norwegian University of Science and Technology}

Method 4 was from Ahmed Mohammed and Marius Pedersen from Norwegian University of Science and Technology, Norway and was based on their network, StreoScenNet \cite{mohammed2019streoscennet}. Their proposed architecture consists of two ResNet \cite{he2015deep} encoder blocks and stacked convolutional decoder network connected with a novel sum-skip connection. This architecture was designed to prevent performance loss from domain shift, by pre-training one encoder with ImageNet. The input to the network is a pair of left and right frames, one to each encoder, and the output is a single mask of the segmented regions for the left frame. It is trained end-to-end and the segmentation is achieved without the need of any pre- or post-processing. The source code can be downloaded from \url{https://github.com/ahme0307/streoscene}. 

\subsection{Indian Institute of Technology, Madras}

Method 5 was from Avinash Kori, Varghese Alex and Ganapathy Krishnamurthi at the Indian Institute of Technhology, Madras, India. They addressed the problem by splitting the task into segmentation of robotic tools and organs and training separate networks for each task. They used a 77 layered fully convolutional dense network architecture and trained the network with a combination of weighted cross entropy and dice loss using the Adam optimizer with a learning rate of 0.001 and a decay rate of 0.1. As a post-processing step, conditional random field (CRF) inference was performed on the logits to reduce false positives. 

\subsection{Ostbayerische Technische Hochschule Regensburg}

Method 6 was from David Rauber, Robert Mendel, Christoph Palm  at Ostbayerische Technische Hochschule Regensburg, Germany. They trained DeepLab V3+, initialized with ResNet-50 pre-trained on ImageNet, using leave-one-out cross-validation. They trained 40 epochs with the 2017 MICCAI data using a learning rate of $1\times10^{-5}$ and the Adam optimizer, followed by 20 epochs using the 2018 data with a learning rate of $1\times10^{-4}$ and then 20 epochs using a learning rate of $1\times10^{-5}$, both with the Adam optimizer. They performed data augmentation by splitting the image into 5 sub-images of size $640 \times 512$ using the 4 corners and center of the image. They then applied random scale in the range $[0.5, 1.5]$, random rotation in the range $[-90, 90]$, random brightness in the range $[-0.2, 0.2]$, random horizontal/vertical flips as well as normalization.

\subsection{University College London (UCL)}

Method 7 was from Sophia Bano and Danail Stoyanov at University College London, UK. They used a global convolutional network (GCN) \cite{peng2017large} with a $11\times11$ kernel size and pretrained ResNet 152 backbone for simultaneous classification and localization. GCN generates semantic score maps by using symmetric and separable large filters to reduce model parameters and computation cost. A boundary refinement improves localization performance near the object boundaries. Data augmentation applied during training was brightness modification in the range $[0.4, 1.4]$, rotation in the range $[-30, 30]$ and random valid image crops of size $512 \times 512$.

\subsection{National Center for Tumor Diseases (NCT)}

Method 8 was from Stefan Leger, Sebastian Bodenstedt and Stephanie Speidel at the National Center for Tumor Diseases, Dresden, Germany. They used a U-Net with a VGG 16 encoder, training a single model for each segmentation class. Pre-training for the instrument classes was performed using the EndoVis 2017 instrument segmentation challenge \cite{allan_endovis_2017} applying augmentations of scaling, rotation, flipping, brightness and contrast.

\subsection{IRCAD}

Method 9 was from Guinther Saibro of IRCAD, France. They also trained a DeepLab V3+ model for this task. The data augmentations they applied was affine transformations, noise and color transformation, applying 20 augmentations for each image.

\subsection{National Taiwan University}

Method 10 was from Chi-Sheng (Daniel) Shih and Hsun-An Chiang at the National Taiwan University, Taiwan. Their method is based on PSPNet \cite{zhao2016pyramid} and they divide the classes of the challenge into semantically similar groups. They first divide into artificial objects and anatomical objects, these classes are then divided into surgical instruments and other artificial objects, and kidney, small intestine and other anatomical objects. From there, these grouped classes were subdivided into the true labels. They also provided a data augmentation technique called collage, which combines parts from different images to form a new training sample.

\subsection{Yale University (MEDYI)}

Method 11 was from Juntang Zhuang and Junlin Yang at Yale University, USA. They trained a modified U-Net with a ResNet-101 backbone. Training was performed using SGD with initial learning rate of 0.01, and it decays at epoch 20 and epoch 30 by 0.1. Focal loss was combined with weighted cross entropy loss. An ensemble of 4 models were trained and aggregated with majority voting to obtain the final prediction.

\subsection{ODS.ai}

The Open Data Science team submission was from Vladimir Iglovikov and Anton Dobrenkii. They trained a WideResnet38 encoder using an InPlace activated batch norm (ABN) \cite{bul_inplace_2017}, with DeepLab V3 \cite{chen_encoder_2018} as the decoder. They used focal loss with $\gamma=2$ using SGD with momentum and a learning rate of 0.035 for 100 epochs. To preprocess the images, they cropped to $712\times712$ and performed random spatial and photometric augmentations.

\subsection{Rediminds Inc.}

The submission from Rediminds Inc., USA was from Madhu Reddiboina and Anubhav Reddy. They trained 3 U-Net models to complete the task. A single model was designed to predict background, instrument and kidney as superclasses, this was trained with a learning rate of 0.001 and a decay factor of 0.5 using the Adam optimizer and a cross entropy loss. Another model was trained to predict the instrument shaft, clasper and wrist along with the suturing clips. This model was trained in the same way as the first except the Jaccard index was used as the loss. A final model predicted the instrument wrist and clamp trained with the Dice loss. These models predicted a final label for each pixel as an ensemble.  

\subsection{Johns Hopkins University (JHU)}

The submission from Johns Hopkins University, USA was from Xingtong Liu, Cong Gao and Mathias Unberath. They trained a Pix2Pix model \cite{imagetoimage_isola_2016} to perform the segmentation, with a U-Net as the generator. They used class weighted cross entropy, L1 loss and adversarial loss to train the model as well as pretraining it as a re-colorization network. Data augmentation was performed by resizing the images to $256 \times 256$ as well as performing standard flips, crops, scalings, rotations and translations along with HSV jitter. To generate predictions they simulated new data by apply random spatial and photometric perturbations to each input sample, then reversing those transformations on the prediction and perform ensemble voting on the multiple outputs. 

\subsection{Daegu Gyeongbuk Institute of Science and Technology (DGIST)}

The submission the Medical Image and Signal Processing Lab at DGIST, South Korea was from Myeonghyeon Kim, Chanho Kim, Chaewon Kim, Hyejin Kim, Gyeongmin Lee, Ihsan Ullah, Miguel Luna and Sang Hyun Park. They trained a U-Net with 5 levels of 2 $3\times3$ convolutional layers, ReLu activation and 1 max pooling layer each. The loss function of the model was defined with categorical cross entropy and the model was trained using Adam optimizer with the learning rate: $0.001/(1 + 0.002 \times epoch)$. Each times was downsampled to $384\times480$ and then standardized. The model was trained for 600 epochs.

\subsection{National University of Singapore (NUS)}

The submission from NUS, Singapore was from Mobarakol Islam. No additional details were supplied about the submission. 

\subsection{Team Banana}

The submission for Team Banana was not accompanied by any data about the team or details about the submission.

%% file: tex/results.tex
\section{Results}

The methods were evaluated using the mean intersection over union (IoU) metric, a current standard for assessing segmentation scores in computer vision literature \cite{lin_coco_2015}. The IoU for a single class is defined as 
\begin{equation}
IOU = TP/(TP+FP+FN)
\end{equation}
where TP is the number of true positive predictions for a class label, FP is the number of false positives and FN is the number of false negatives. To compute the mean IoU we use the arithmetic mean of the IoU for all classes that are present in a given frame. If we are considering a set of classes and none are present in the frame, we discount the frame from the calculation. We compute this score for each frame and average over all frames to get a per-dataset score. When computing overall scores we weight each score by the size of the dataset. 

\input{tex/test_dataset_1}
\input{tex/test_dataset_2}
\input{tex/test_dataset_3}
\input{tex/test_dataset_4}

\begin{table*}
\centering
\small
\begin{tabular}{ 
    p{0.11\textwidth} | %l 
    >{\centering}p{0.08\textwidth} | 
    >{\centering}p{0.06\textwidth} | 
    >{\centering}p{0.07\textwidth} | 
    >{\centering}p{0.07\textwidth} | 
    >{\centering}p{0.07\textwidth} | 
    >{\centering}p{0.04\textwidth} | 
    >{\centering}p{0.04\textwidth} | 
    >{\centering}p{0.05\textwidth} | 
    >{\centering}p{0.05\textwidth} | 
    >{\centering}p{0.03\textwidth} | 
    p{0.05\textwidth}
}
 & Parenchyma & Covered Kidney & Instrument Shaft & Instrument Clasper & Instrument Wrist & Thread & Needle & US Probe & Intestine & Clips & Overall \\
\hline
Team Banana     & 0.399 & 0.246 & 0.728 & 0.354 & 0.403 & 0.100 & 0.000 & 0.000 & 0.321 & 0.407 & 0.435 \\
Satoshi Kondo   & 0.479 & \textbf{0.464} & 0.772 & 0.404 & 0.396 & 0.283 & 0.000 & 0.000 & 0.529 & 0.609 & 0.495 \\
NUS             & 0.337 & 0.180 & 0.665 & 0.360 & 0.384 & 0.006 & 0.000 & 0.000 & 0.364 & 0.015 & 0.394 \\
NCT 1           & 0.387 & 0.177 & 0.729 & 0.353 & 0.380 & 0.008 & 0.000 & 0.235 & 0.360 & 0.079 & 0.411 \\
Digital Surgery & 0.646 & 0.208 & 0.822 & 0.580 & 0.553 & 0.061 & 0.000 & 0.199 & 0.372 & 0.772 & 0.579 \\
Fan Voyage      & 0.626 & 0.326 & 0.819 & 0.529 & 0.512 & 0.065 & 0.000 & 0.256 & 0.503 & 0.703 & 0.570 \\
IIT Madras      & 0.272 & 0.053 & 0.596 & 0.317 & 0.293 & 0.031 & 0.000 & 0.083 & 0.198 & 0.102 & 0.311 \\
OTH Regensburg  & \textbf{0.654} & 0.282 & \textbf{0.845} & \textbf{0.658} & \textbf{0.610} & \textbf{0.478} & 0.014 & 0.282 & 0.337 & \textbf{0.849} & \textbf{0.621} \\
UCL             & 0.449 & 0.315 & 0.763 & 0.401 & 0.457 & 0.106 & 0.000 & 0.104 & 0.313 & 0.788 & 0.485 \\
NCT 2           & 0.618 & 0.293 & 0.777 & 0.622 & 0.560 & 0.330 & \textbf{0.019} & \textbf{0.335} & \textbf{0.562} & 0.729 & 0.585 \\
IRCAD           & 0.580 & 0.144 & 0.842 & 0.635 & 0.595 & 0.315 & 0.000 & 0.224 & 0.103 & 0.505 & 0.573 \\
Nat. Taiwan U.  & 0.495 & 0.102 & 0.746 & 0.367 & 0.414 & 0.133 & 0.000 & 0.004 & 0.236 & 0.680 & 0.439 \\
MEDYI           & 0.500 & 0.276 & 0.746 & 0.420 & 0.496 & 0.002 & 0.000 & 0.011 & 0.270 & 0.624 & 0.486 \\
DGIST           & 0.444 & 0.084 & 0.617 & 0.373 & 0.408 & 0.030 & 0.000 & 0.098 & 0.240 & 0.461 & 0.398 \\
UNC             & 0.614 & 0.363 & 0.839 & 0.622 & 0.542 & 0.475 & 0.000 & 0.305 & 0.545 & 0.771 & 0.607 \\
ODS.ai          & 0.483 & 0.170 & 0.815 & 0.599 & 0.569 & 0.223 & 0.012 & 0.176 & 0.490 & 0.778 & 0.546 \\
JHU             & 0.533 & 0.349 & 0.747 & 0.464 & 0.387 & 0.129 & 0.001 & 0.000 & 0.248 & 0.396 & 0.491 \\
Rediminds Inc.  & 0.089 & 0.000 & 0.409 & 0.206 & 0.276 & 0.000 & 0.000 & 0.000 & 0.000 & 0.380 & 0.192 \\
\hline
Average & 0.479 & 0.224 & 0.738 & 0.460 & 0.457 & 0.154 & 0.003 & 0.128 & 0.332 & 0.536 & 0.478
\end{tabular}
\caption{\label{tab:dataset_overall_results} The overall numerical results achieved by averaging across the 4 test datasets. The highest scoring method is shown in bold. 6 classes are won by the team from OTH Regensburg, 1 by Satoshi Kondo and 3 by the team from NCT.}
\end{table*}

%% file: tex/test_dataset_1.tex
\subsection{Test Dataset 1}

Test dataset 1 contains a single zoomed out sequence panning across the liver and stomach before arriving at the kidney parenchyma. The kidney has the fascia and perirenal fat removed and the surgeon dissects a single small piece of kidney tissue with a monopolar scissor instrument. The resected cavity is then sutured up using 2 large needle driver instruments. The numerical results for this dataset are displayed in Table \ref{tab:dataset_1_results} and show that 3 teams scored above 0.9 IoU for the kidney class and an average score of 0.674, just below the score on the kidney for the best performing dataset, which was dataset 3. Most of the sequence focused on a single close up view of an exposed parenchyma, which is far easier to recognize compared with the fascia covered kidney. Most teams struggled at the start which had a exploratory sequence and camera view obstruction by the cannula. 

\begin{table*}
\centering
\small
\begin{tabular}{ 
    p{0.13\textwidth} | %l 
    >{\centering}p{0.08\textwidth} | 
    >{\centering}p{0.07\textwidth} | 
    >{\centering}p{0.07\textwidth} | 
    >{\centering}p{0.07\textwidth} | 
    >{\centering}p{0.05\textwidth} | 
    >{\centering}p{0.05\textwidth} | 
    >{\centering}p{0.05\textwidth} | 
    >{\centering}p{0.05\textwidth} | 
    >{\centering}p{0.05\textwidth} | 
    p{0.05\textwidth}
}
  & Parenchyma & Instrument Shaft & Instrument Clasper & Instrument Wrist & Thread & Needle & US Probe & Intestine & Clips & Overall \\
\hline
Team Banana     & 0.577 & 0.710 & 0.220 & 0.347 & 0.100 & 0.000 & 0.000 & 0.107 & 0.407 & 0.386 \\
Satoshi Kondo   & 0.666 & 0.763 & 0.280 & 0.384 & 0.283 & 0.000 & 0.000 & 0.265 & 0.609 & 0.447 \\
NUS             & 0.519 & 0.659 & 0.442 & 0.374 & 0.006 & 0.000 & 0.000 & 0.055 & 0.015 & 0.403 \\
NCT 1           & 0.508 & 0.689 & 0.218 & 0.331 & 0.008 & 0.000 & 0.213 & 0.148 & 0.079 & 0.359 \\
Digital Surgery & 0.904 & 0.820 & 0.685 & 0.593 & 0.061 & 0.000 & 0.302 & 0.166 & 0.772 & 0.636 \\
Fan Voyage      & 0.882 & 0.842 & 0.584 & 0.509 & 0.065 & 0.000 & 0.338 & 0.234 & 0.703 & 0.601 \\
IIT Madras      & 0.607 & 0.597 & 0.181 & 0.241 & 0.031 & 0.000 & 0.126 & 0.177 & 0.102 & 0.331 \\
OTH Regensburg  & 0.810 & 0.847 & \textbf{0.775} & \textbf{0.668} & \textbf{0.478} & 0.014 & 0.389 & 0.231 & \textbf{0.849} & \textbf{0.691} \\
UCL             & 0.514 & 0.711 & 0.306 & 0.371 & 0.106 & 0.000 & 0.134 & 0.098 & 0.788 & 0.425 \\
NCT 2           & \textbf{0.906} & 0.755 & 0.718 & 0.603 & 0.330 & \textbf{0.019} & 0.442 & \textbf{0.270} & 0.729 & 0.658 \\
IRCAD           & 0.904 & \textbf{0.849} & 0.739 & 0.642 & 0.315 & 0.000 & \textbf{0.447} & 0.138 & 0.505 & 0.688 \\
Nat. Taiwan U.  & 0.801 & 0.742 & 0.452 & 0.375 & 0.133 & 0.000 & 0.007 & 0.030 & 0.680 & 0.505 \\
MEDYI           & 0.495 & 0.750 & 0.474 & 0.497 & 0.002 & 0.000 & 0.020 & 0.240 & 0.624 & 0.469 \\
DGIST           & 0.759 & 0.635 & 0.412 & 0.409 & 0.030 & 0.000 & 0.192 & 0.078 & 0.461 & 0.474 \\
UNC             & 0.866 & 0.834 & 0.703 & 0.559 & 0.475 & 0.000 & 0.428 & 0.224 & 0.771 & 0.663 \\
ODS.ai          & 0.572 & 0.826 & 0.690 & 0.585 & 0.223 & 0.012 & 0.244 & 0.202 & 0.778 & 0.585 \\
JHU             & 0.681 & 0.764 & 0.459 & 0.364 & 0.129 & 0.001 & 0.000 & 0.078 & 0.396 & 0.470 \\
Rediminds Inc.  & 0.158 & 0.375 & 0.154 & 0.259 & 0.000 & 0.000 & 0.000 & 0.000 & 0.380 & 0.212 \\
\hline
Average & 0.674 & 0.731 & 0.472 & 0.451 & 0.154 & 0.003 & 0.182 & 0.152 & 0.536 & 0.500
\end{tabular}
\caption{\label{tab:dataset_1_results} The numerical results for the test dataset 1. The highest scoring method is shown in bold. 4 classes were won by the team from OTH Regensburg, 2 classes were won by the team from NCT and 2 by the team from IRCAD.}
\end{table*}

\begin{figure*}
\captionsetup[subfigure]{labelformat=empty}
\begin{subfigure}[b]{0.23\textwidth}
\includegraphics[width=\textwidth]{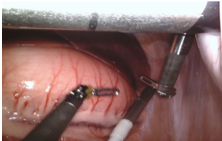}
\caption{Frame 6}
\end{subfigure}
\hfill
\begin{subfigure}[b]{0.23\textwidth}
\includegraphics[width=\textwidth]{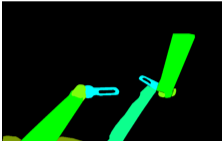}
\caption{Ground Truth}
\end{subfigure}
\hfill
\begin{subfigure}[b]{0.23\textwidth}
\includegraphics[width=\textwidth]{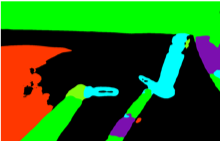}
\caption{IRCAD}
\end{subfigure}
\hfill
\begin{subfigure}[b]{0.23\textwidth}
\includegraphics[width=\textwidth]{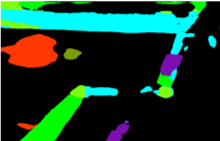}
\caption{UCL}
\end{subfigure}
\\
\begin{subfigure}[b]{0.23\textwidth}
\includegraphics[width=\textwidth]{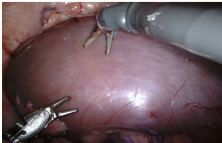}
\caption{Frame 76}
\end{subfigure}
\hfill
\begin{subfigure}[b]{0.23\textwidth}
\includegraphics[width=\textwidth]{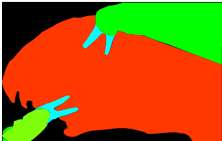}
\caption{Ground Truth}
\end{subfigure}
\hfill
\begin{subfigure}[b]{0.23\textwidth}
\includegraphics[width=\textwidth]{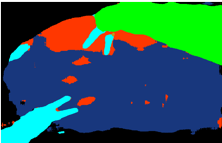}
\caption{Satoshi Kondo}
\end{subfigure}
\hfill
\begin{subfigure}[b]{0.23\textwidth}
\includegraphics[width=\textwidth]{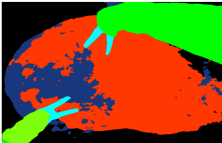}
\caption{OTH Regensburg}
\end{subfigure}
\\\begin{subfigure}[b]{0.23\textwidth}
\includegraphics[width=\textwidth]{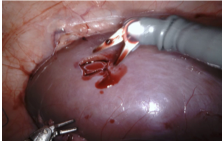}
\caption{Frame 122}
\end{subfigure}
\hfill
\begin{subfigure}[b]{0.23\textwidth}
\includegraphics[width=\textwidth]{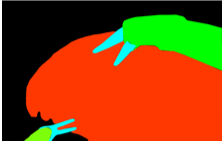}
\caption{Ground Truth}
\end{subfigure}
\hfill
\begin{subfigure}[b]{0.23\textwidth}
\includegraphics[width=\textwidth]{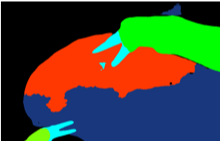}
\caption{ODS.ai}
\end{subfigure}
\hfill
\begin{subfigure}[b]{0.23\textwidth}
\includegraphics[width=\textwidth]{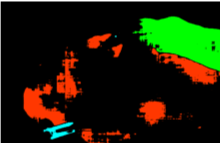}
\caption{Rediminds Inc.}
\end{subfigure}
\\\begin{subfigure}[b]{0.23\textwidth}
\includegraphics[width=\textwidth]{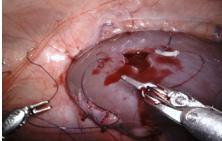}
\caption{Frame 276}
\end{subfigure}
\hfill
\begin{subfigure}[b]{0.23\textwidth}
\includegraphics[width=\textwidth]{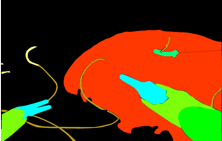}
\caption{Ground Truth}
\end{subfigure}
\hfill
\begin{subfigure}[b]{0.23\textwidth}
\includegraphics[width=\textwidth]{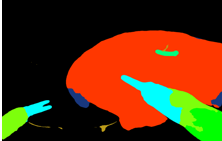}
\caption{Digital Surgery}
\end{subfigure}
\hfill
\begin{subfigure}[b]{0.23\textwidth}
\includegraphics[width=\textwidth]{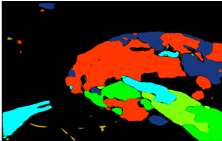}
\caption{Team Banana}
\end{subfigure}
\caption{\label{fig:test_dataset_1_qualitative}Qualitative results for test dataset 1 showing frames 6, 76, 122 and 176 alongside the ground truth images and submission images from randomly chosen teams.}
\end{figure*}

%% file: tex/test_dataset_2.tex
\subsection{Test Dataset 2}

\begin{figure*}
\captionsetup[subfigure]{labelformat=empty}
\begin{subfigure}[b]{0.23\textwidth}
\includegraphics[width=\textwidth]{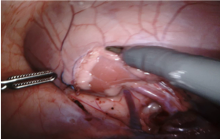}
\caption{Frame 6}
\end{subfigure}
\hfill
\begin{subfigure}[b]{0.23\textwidth}
\includegraphics[width=\textwidth]{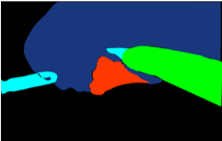}
\caption{Ground Truth}
\end{subfigure}
\hfill
\begin{subfigure}[b]{0.23\textwidth}
\includegraphics[width=\textwidth]{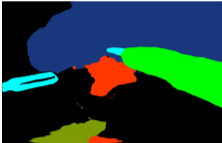}
\caption{NUS}
\end{subfigure}
\hfill
\begin{subfigure}[b]{0.23\textwidth}
\includegraphics[width=\textwidth]{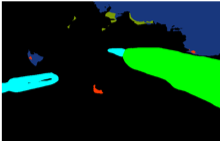}
\caption{National Taiwan University}
\end{subfigure}
\\
\begin{subfigure}[b]{0.23\textwidth}
\includegraphics[width=\textwidth]{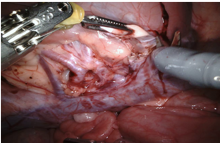}
\caption{Frame 90}
\end{subfigure}
\hfill
\begin{subfigure}[b]{0.23\textwidth}
\includegraphics[width=\textwidth]{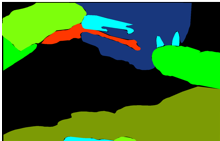}
\caption{Ground Truth}
\end{subfigure}
\hfill
\begin{subfigure}[b]{0.23\textwidth}
\includegraphics[width=\textwidth]{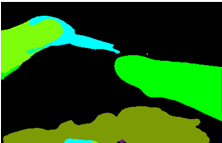}
\caption{NCT 2}
\end{subfigure}
\hfill
\begin{subfigure}[b]{0.23\textwidth}
\includegraphics[width=\textwidth]{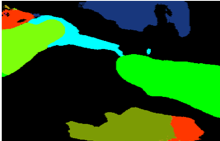}
\caption{JHU}
\end{subfigure}
\\\begin{subfigure}[b]{0.23\textwidth}
\includegraphics[width=\textwidth]{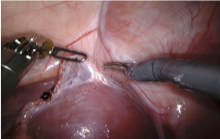}
\caption{Frame 160}
\end{subfigure}
\hfill
\begin{subfigure}[b]{0.23\textwidth}
\includegraphics[width=\textwidth]{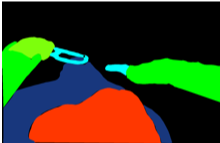}
\caption{Ground Truth}
\end{subfigure}
\hfill
\begin{subfigure}[b]{0.23\textwidth}
\includegraphics[width=\textwidth]{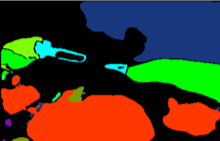}
\caption{NCT 1}
\end{subfigure}
\hfill
\begin{subfigure}[b]{0.23\textwidth}
\includegraphics[width=\textwidth]{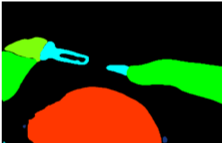}
\caption{Fan Voyage}
\end{subfigure}
\\\begin{subfigure}[b]{0.23\textwidth}
\includegraphics[width=\textwidth]{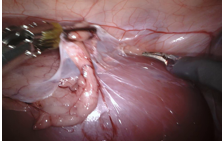}
\caption{Frame 200}
\end{subfigure}
\hfill
\begin{subfigure}[b]{0.23\textwidth}
\includegraphics[width=\textwidth]{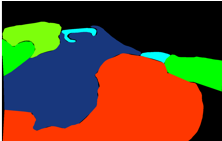}
\caption{Ground Truth}
\end{subfigure}
\hfill
\begin{subfigure}[b]{0.23\textwidth}
\includegraphics[width=\textwidth]{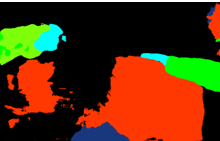}
\caption{DGIST}
\end{subfigure}
\hfill
\begin{subfigure}[b]{0.23\textwidth}
\includegraphics[width=\textwidth]{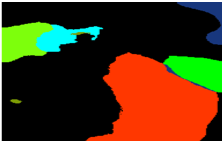}
\caption{IIT Madras}
\end{subfigure}
\caption{\label{fig:test_dataset_2_qualitative}Qualitative results for test dataset 2 showing frames 6, 90, 160 and 200 alongside the ground truth images and submission images from randomly chosen teams.}
\end{figure*}

Test dataset 2 maintains a close up view of a kidney which is initially covered by renal fascia and perirenal fat and this is gradually removed using a monopolar curved scissor and fenestrated bipolar forceps instrument. The numerical results are shown in Table \ref{tab:dataset_2_results}. The parenchyma is segmented more accurately than the `covered kidney' label with more than $2x$ the average mean IoU. The qualitative results in Figure \ref{fig:test_dataset_2_qualitative} illustrate that this class is often mistaken for background and vice-versa.

\begin{table*}
\centering
\small
%\begin{tabular}{ l |  c | c | c | c | c | c }
\begin{tabular}{ p{0.13\textwidth} |  >{\centering}p{0.1\textwidth} | >{\centering}p{0.1\textwidth} | >{\centering}p{0.1\textwidth} | >{\centering}p{0.1\textwidth} | >{\centering}p{0.1\textwidth}  | >{\centering}p{0.1\textwidth} | p{0.1\textwidth}}
& Parenchyma & Covered Kidney & Instrument Shaft & Instrument Clasper & Instrument Wrist & Intestine & Overall \\
\hline
Team Banana     & 0.295 & 0.135 & 0.875 & 0.431 & 0.429 & 0.227 & 0.426 \\
Satoshi Kondo   &  0.358 & 0.363 & 0.885 & 0.479 & 0.453 & 0.444 & 0.505 \\
NUS             &  0.271 & 0.156 & 0.850 & 0.418 & 0.387 & 0.212 & 0.407 \\
NCT 1           &  0.394 & 0.209 & 0.841 & 0.400 & 0.379 & 0.467 & 0.448 \\
Digital Surgery &  0.610 & 0.277 & 0.923 & 0.464 & 0.500 & 0.336 & 0.549 \\
Fan Voyage      &  0.542 & 0.289 & 0.915 & 0.477 & 0.491 & 0.471 & 0.542 \\
IIT Madras      &  0.293 & 0.068 & 0.770 & 0.420 & 0.312 & 0.190 & 0.370 \\
OTH Regensburg  &  \textbf{0.667} & 0.296 & 0.917 & \textbf{0.512} & 0.525 & 0.381 & 0.575 \\
UCL             &  0.384 & 0.257 & 0.885 & 0.467 & 0.482 & 0.299 & 0.487 \\
NCT 2           &  0.572 & 0.304 & 0.837 & 0.511 & 0.514 & \textbf{0.659} & 0.555 \\
IRCAD           &  0.523 & 0.246 & \textbf{0.925} & 0.515 & \textbf{0.531} & 0.095 & 0.529 \\
Nat. Taiwan U.  &  0.356 & 0.116 & 0.884 & 0.375 & 0.400 & 0.116 & 0.416 \\
MEDYI           &  0.619 & 0.253 & 0.886 & 0.439 & 0.463 & 0.292 & 0.524 \\
DGIST           &  0.447 & 0.093 & 0.844 & 0.389 & 0.374 & 0.077 & 0.418 \\
UNC             &  0.635 & \textbf{0.375} & 0.916 & 0.498 & 0.460 & 0.590 & \textbf{0.578} \\
ODS.ai          &  0.453 & 0.180 & 0.901 & 0.487 & 0.494 & 0.628 & 0.508 \\
JHU             &  0.531 & 0.270 & 0.866 & 0.459 & 0.387 & 0.192 & 0.490 \\
Rediminds Inc.  &  0.138 & 0.000 & 0.646 & 0.371 & 0.330 & 0.000 & 0.284 \\
\hline
Average & 0.449 & 0.216 & 0.865 & 0.451 & 0.440 & 0.315 & 0.478
\end{tabular}
\caption{\label{tab:dataset_2_results} The numerical results for the test dataset 2. The highest scoring method is shown in bold. 2 classes were won by the team from OTH Regensburg, 2 by IRCAD, and 1 by NCT and 1 by UNC.}
\end{table*}

%% file: tex/test_dataset_3.tex
\subsection{Test Dataset 3}

Similarly to test dataset 1, test dataset 3 has a close-up sequence of the parenchyma which has been exposed by the removal of the fascia and fat. The instruments in the sequence are a large needle driver and a large needle driver. The camera moves from the center of the parenchyma to focus on a hilar exposure. is looks like a straightforward sequence with many frames showing a complete close-up view of the kidney however the scores are quite poor, mostly likely due to the heavily covered surface of the kidney. 

\begin{table*}
\centering
\small
%\begin{tabular}{ l |  c | c | c | c | c | c }
\begin{tabular}{ m{0.18\textwidth} |  >{\centering}m{0.1\textwidth} | >{\centering}m{0.1\textwidth} | >{\centering}m{0.1\textwidth} | >{\centering}m{0.1\textwidth}  | >{\centering}m{0.1\textwidth} | m{0.1\textwidth}}
  & Parenchyma & Instrument Shaft & Instrument Clasper & Instrument Wrist & Intestine & Overall \\
\hline
Team Banana     & 0.641 & 0.882 & 0.503 & 0.481 & 0.365 & 0.622 \\
Satoshi Kondo   & 0.836 & 0.847 & 0.495 & 0.441 & 0.828 & 0.658 \\
NUS             & 0.510 & 0.798 & 0.398 & 0.476 & 0.688 & 0.546 \\
NCT 1           & 0.539 & 0.839 & 0.520 & 0.423 & 0.334 & 0.573 \\
Digital Surgery & 0.908 & 0.916 & 0.736 & 0.672 & 0.316 & 0.806 \\
Fan Voyage      & 0.912 & 0.912 & 0.645 & 0.607 & 0.715 & 0.771 \\
IIT Madras      & 0.164 & 0.646 & 0.435 & 0.354 & 0.075 & 0.392 \\
OTH Regensburg  & 0.875 & \textbf{0.922} & \textbf{0.813} & \textbf{0.733} & 0.153 & \textbf{0.829} \\
UCL             & 0.789 & 0.889 & 0.547 & 0.572 & 0.321 & 0.699 \\
NCT 2           & \textbf{0.888} & 0.785 & 0.763 & 0.625 & 0.676 & 0.765 \\
IRCAD           & 0.777 & 0.927 & 0.767 & 0.720 & 0.102 & 0.790 \\
Nat. Taiwan U.  & 0.812 & 0.820 & 0.456 & 0.555 & 0.411 & 0.657 \\
MEDYI           & 0.786 & 0.818 & 0.478 & 0.648 & 0.103 & 0.679 \\
DGIST           & 0.567 & 0.721 & 0.451 & 0.548 & 0.559 & 0.569 \\
UNC             & 0.871 & 0.912 & 0.794 & 0.680 & \textbf{0.763} & 0.814 \\
ODS.ai          & 0.841 & 0.913 & 0.742 & 0.711 & 0.509 & 0.799 \\
JHU             & 0.806 & 0.804 & 0.561 & 0.450 & 0.175 & 0.653 \\
Rediminds Inc.  & 0.050 & 0.416 & 0.128 & 0.294 & 0.000 & 0.220 \\
\hline
Average & 0.698 & 0.820 & 0.568 & 0.555 & 0.394 & 0.658
\end{tabular}
\caption{\label{tab:dataset_3_results} The numerical results for the test dataset 3. The highest scoring method is shown in bold. 4 classes were won by team from OTH Regensburg, 1 by the team from NCT and 1 by the team from UNC.}
\end{table*}

\begin{figure*}
\captionsetup[subfigure]{labelformat=empty}
\begin{subfigure}[b]{0.23\textwidth}
\includegraphics[width=\textwidth]{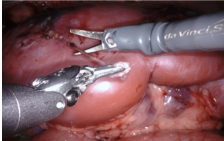}
\caption{Frame 17}
\end{subfigure}
\hfill
\begin{subfigure}[b]{0.23\textwidth}
\includegraphics[width=\textwidth]{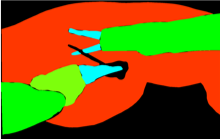}
\caption{Ground Truth}
\end{subfigure}
\hfill
\begin{subfigure}[b]{0.23\textwidth}
\includegraphics[width=\textwidth]{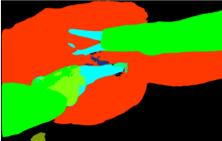}
\caption{MEDYI}
\end{subfigure}
\hfill
\begin{subfigure}[b]{0.23\textwidth}
\includegraphics[width=\textwidth]{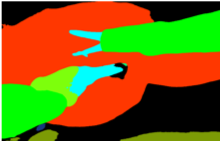}
\caption{ODS.ai}
\end{subfigure}
\\
\begin{subfigure}[b]{0.23\textwidth}
\includegraphics[width=\textwidth]{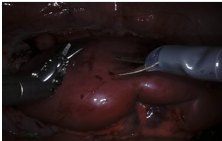}
\caption{Frame 100}
\end{subfigure}
\hfill
\begin{subfigure}[b]{0.23\textwidth}
\includegraphics[width=\textwidth]{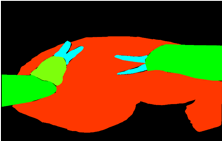}
\caption{Ground Truth}
\end{subfigure}
\hfill
\begin{subfigure}[b]{0.23\textwidth}
\includegraphics[width=\textwidth]{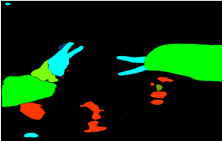}
\caption{Team Banana}
\end{subfigure}
\hfill
\begin{subfigure}[b]{0.23\textwidth}
\includegraphics[width=\textwidth]{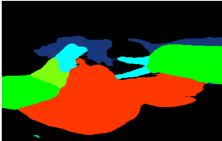}
\caption{UNC}
\end{subfigure}
\\\begin{subfigure}[b]{0.23\textwidth}
\includegraphics[width=\textwidth]{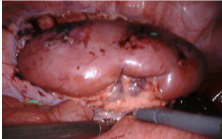}
\caption{Frame 126}
\end{subfigure}
\hfill
\begin{subfigure}[b]{0.23\textwidth}
\includegraphics[width=\textwidth]{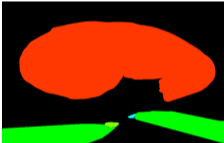}
\caption{Ground Truth}
\end{subfigure}
\hfill
\begin{subfigure}[b]{0.23\textwidth}
\includegraphics[width=\textwidth]{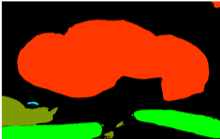}
\caption{National Taiwan University}
\end{subfigure}
\hfill
\begin{subfigure}[b]{0.23\textwidth}
\includegraphics[width=\textwidth]{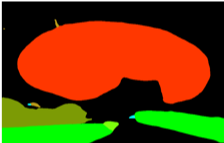}
\caption{IRCAD}
\end{subfigure}
\\\begin{subfigure}[b]{0.23\textwidth}
\includegraphics[width=\textwidth]{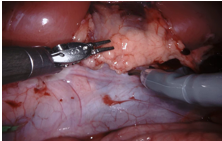}
\caption{Frame 177}
\end{subfigure}
\hfill
\begin{subfigure}[b]{0.23\textwidth}
\includegraphics[width=\textwidth]{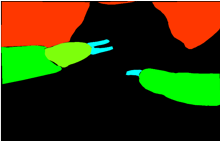}
\caption{Ground Truth}
\end{subfigure}
\hfill
\begin{subfigure}[b]{0.23\textwidth}
\includegraphics[width=\textwidth]{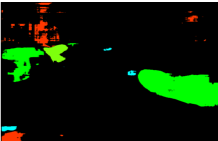}
\caption{Team Rediminds}
\end{subfigure}
\hfill
\begin{subfigure}[b]{0.23\textwidth}
\includegraphics[width=\textwidth]{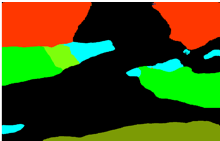}
\caption{Satoshi Kondo}
\end{subfigure}
\caption{\label{fig:test_dataset_3_qualitative}Qualitative results for test dataset 3 showing frames 17, 100, 126 and 177 alongside the ground truth images and submission images from randomly chosen teams.}
\end{figure*}

%% file: tex/test_dataset_4.tex
\subsection{Test Dataset 4}

\begin{figure*}
\captionsetup[subfigure]{labelformat=empty}
\begin{subfigure}[b]{0.23\textwidth}
\includegraphics[width=\textwidth]{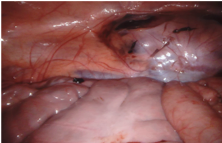}
\caption{Frame 10}
\end{subfigure}
\hfill
\begin{subfigure}[b]{0.23\textwidth}
\includegraphics[width=\textwidth]{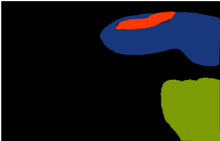}
\caption{Ground Truth}
\end{subfigure}
\hfill
\begin{subfigure}[b]{0.23\textwidth}
\includegraphics[width=\textwidth]{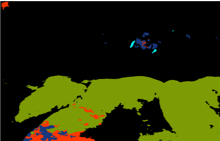}
\caption{OTH Regensburg}
\end{subfigure}
\hfill
\begin{subfigure}[b]{0.23\textwidth}
\includegraphics[width=\textwidth]{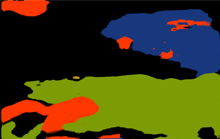}
\caption{NUS}
\end{subfigure}
\\
\begin{subfigure}[b]{0.23\textwidth}
\includegraphics[width=\textwidth]{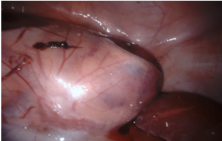}
\caption{Frame 55}
\end{subfigure}
\hfill
\begin{subfigure}[b]{0.23\textwidth}
\includegraphics[width=\textwidth]{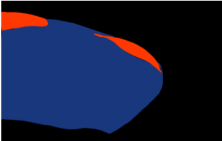}
\caption{Ground Truth}
\end{subfigure}
\hfill
\begin{subfigure}[b]{0.23\textwidth}
\includegraphics[width=\textwidth]{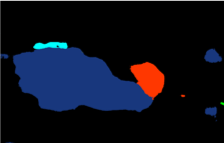}
\caption{UCL}
\end{subfigure}
\hfill
\begin{subfigure}[b]{0.23\textwidth}
\includegraphics[width=\textwidth]{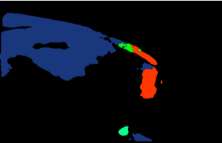}
\caption{NCT 2}
\end{subfigure}
\\\begin{subfigure}[b]{0.23\textwidth}
\includegraphics[width=\textwidth]{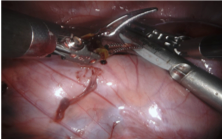}
\caption{Frame 125}
\end{subfigure}
\hfill
\begin{subfigure}[b]{0.23\textwidth}
\includegraphics[width=\textwidth]{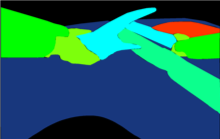}
\caption{Ground Truth}
\end{subfigure}
\hfill
\begin{subfigure}[b]{0.23\textwidth}
\includegraphics[width=\textwidth]{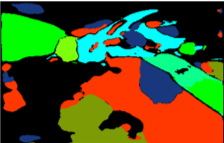}
\caption{NCT 1}
\end{subfigure}
\hfill
\begin{subfigure}[b]{0.23\textwidth}
\includegraphics[width=\textwidth]{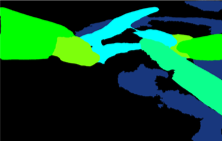}
\caption{JHU}
\end{subfigure}
\\\begin{subfigure}[b]{0.23\textwidth}
\includegraphics[width=\textwidth]{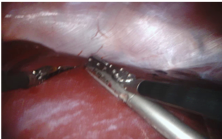}
\caption{Frame 237}
\end{subfigure}
\hfill
\begin{subfigure}[b]{0.23\textwidth}
\includegraphics[width=\textwidth]{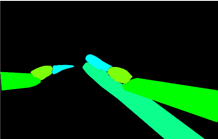}
\caption{Ground Truth}
\end{subfigure}
\hfill
\begin{subfigure}[b]{0.23\textwidth}
\includegraphics[width=\textwidth]{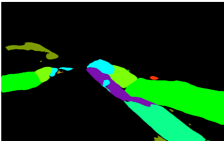}
\caption{Fan Voyage}
\end{subfigure}
\hfill
\begin{subfigure}[b]{0.23\textwidth}
\includegraphics[width=\textwidth]{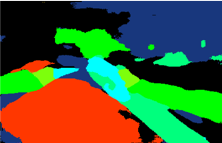}
\caption{UNC}
\end{subfigure}
\caption{\label{fig:test_dataset_4_qualitative}Qualitative results for test dataset 4 showing frames 10, 55, 125 and 237 alongside the ground truth images and submission images from randomly chosen teams.}
\end{figure*}

Test dataset 4 begins from a zoomed out view of a kidney which is occluded by fascia and fat with a large amount of small and large intestine at the bottom of the image. The camera then moves directly over the kidney where a Maryland Bipolar Forceps instrument and a Prograsp Forceps instrument are used to scan the surface with a drop in ultrasound probe. The camera and instruments are then moved to the liver where the US probe is used again. Liver is categorized as background in our dataset. 
The numerical results for test dataset 4 are shown in Table \ref{tab:dataset_4_results} and the qualitative results are shown in Figure \ref{fig:test_dataset_4_qualitative}. The results for the kidney classes were the worst across all of the test datasets, which is expected given that most of the kidney surface is covered.

\begin{table*}
\centering
\small
\begin{tabular}{ 
    p{0.13\textwidth} | %l 
    >{\centering}p{0.08\textwidth} | 
    >{\centering}p{0.08\textwidth} |
    >{\centering}p{0.08\textwidth} | 
    >{\centering}p{0.08\textwidth} | 
    >{\centering}p{0.08\textwidth} | 
    >{\centering}p{0.06\textwidth} | 
    >{\centering}p{0.06\textwidth} | 
    p{0.05\textwidth}
}
& Parenchyma & Covered Kidney & Instrument Shaft & Instrument Clasper & Instrument Wrist  & US Probe & Intestine & Overall \\
\hline
Team Banana     & 0.084 & 0.358 & 0.444 & 0.261 & 0.355 & 0.000 & 0.585 & 0.305 \\
Satoshi Kondo   & 0.056 & \textbf{0.565} & 0.594 & 0.361 & 0.305 & 0.000 & 0.579 & 0.368 \\
NUS             & 0.047 & 0.203 & 0.352 & 0.181 & 0.297 & 0.000 & 0.501 & 0.221 \\
NCT 1           & 0.108 & 0.144 & 0.546 & 0.275 & 0.387 & 0.258 & 0.490 & 0.262 \\
Digital Surgery & 0.161 & 0.138 & 0.631 & 0.436 & 0.446 & 0.096 & 0.671 & 0.324 \\
Fan Voyage      & 0.167 & 0.362 & 0.607 & 0.411 & 0.441 & 0.174 & 0.591 & 0.366 \\
IIT Madras      & 0.026 & 0.038 & 0.370 & 0.232 & 0.267 & 0.040 & 0.351 & 0.149 \\
OTH Regensburg  & \textbf{0.264} & 0.268 & \textbf{0.693} & \textbf{0.532} & \textbf{0.513} & 0.175 & 0.584 & \textbf{0.390} \\
UCL             & 0.108 & 0.374 & 0.567 & 0.284 & 0.401 & 0.075 & 0.533 & 0.330 \\
NCT 2           & 0.106 & 0.282 & 0.730 & 0.495 & 0.499 & 0.229 & 0.644 & 0.362 \\
IRCAD           & 0.116 & 0.042 & 0.666 & 0.520 & 0.487 & 0.000 & 0.078 & 0.285 \\
Nat. Taiwan U.  & 0.011 & 0.087 & 0.537 & 0.186 & 0.324 & 0.000 & 0.386 & 0.176 \\
MEDYI           & 0.100 & 0.299 & 0.531 & 0.288 & 0.376 & 0.002 & 0.445 & 0.272 \\
DGIST           & 0.005 & 0.075 & 0.270 & 0.239 & 0.302 & 0.003 & 0.245 & 0.129 \\
UNC             & 0.084 & 0.351 & 0.693 & 0.491 & 0.469 & \textbf{0.183} & 0.605 & 0.373 \\
ODS.ai          & 0.067 & 0.161 & 0.619 & 0.479 & 0.485 & 0.107 & \textbf{0.619} & 0.293 \\
JHU             & 0.115 & 0.427 & 0.554 & 0.375 & 0.346 & 0.000 & 0.546 & 0.349 \\
Rediminds Inc.  & 0.010 & 0.000 & 0.199 & 0.171 & 0.221 & 0.000 & 0.000 & 0.052 \\
\hline
Average & 0.091 & 0.232 & 0.534 & 0.345 & 0.385 & 0.075 & 0.470 & 0.279
\end{tabular}
\caption{\label{tab:dataset_4_results} The numerical results for the test dataset 4. The highest scoring method is shown in bold. 4 classes are won by the team from OTH Regensburg, 1 by Satoshi Kondo, 1 by UNC and 1 by ODS.ai.}
\end{table*}

%% file: ms.bbl
% Generated by IEEEtran.bst, version: 1.14 (2015/08/26)
\begin{thebibliography}{10}
\providecommand{\url}[1]{#1}
\csname url@samestyle\endcsname
\providecommand{\newblock}{\relax}
\providecommand{\bibinfo}[2]{#2}
\providecommand{\BIBentrySTDinterwordspacing}{\spaceskip=0pt\relax}
\providecommand{\BIBentryALTinterwordstretchfactor}{4}
\providecommand{\BIBentryALTinterwordspacing}{\spaceskip=\fontdimen2\font plus
\BIBentryALTinterwordstretchfactor\fontdimen3\font minus
  \fontdimen4\font\relax}
\providecommand{\BIBforeignlanguage}[2]{{%
\expandafter\ifx\csname l@#1\endcsname\relax
\typeout{** WARNING: IEEEtran.bst: No hyphenation pattern has been}%
\typeout{** loaded for the language `#1'. Using the pattern for}%
\typeout{** the default language instead.}%
\else
\language=\csname l@#1\endcsname
\fi
#2}}
\providecommand{\BIBdecl}{\relax}
\BIBdecl

\bibitem{allan_endovis_2017}
M.~Allan, A.~Shvets, T.~Kurmann, Z.~Zhang, R.~Duggal, Y.-H. Su, N.~Rieke,
  I.~Laina, N.~Kalavakonda, S.~Bodenstedt, L.~Herrera, W.~Li, V.~Iglovikov,
  H.~Luo, J.~Yang, D.~Stoyanov, L.~Maier-Hein, S.~Speidel, and M.~Azizian,
  ``2017 robotic instrument segmentation challenge,'' 2019.

\bibitem{palep_robot_2009}
J.~H. Palep, ``Robotic assisted minimally invasive surgery,'' in \emph{Journal
  of minimal access surgery}, vol.~5, no.~1, 2009, pp. 1--7.

\bibitem{chen2018encoderdecoder}
L.-C. Chen, Y.~Zhu, G.~Papandreou, F.~Schroff, and H.~Adam, ``Encoder-decoder
  with atrous separable convolution for semantic image segmentation,'' 2018.

\bibitem{resnext_xie_2017}
S.~{Xie}, R.~{Girshick}, P.~{Dollár}, Z.~{Tu}, and K.~{He}, ``Aggregated
  residual transformations for deep neural networks,'' in \emph{2017 IEEE
  Conference on Computer Vision and Pattern Recognition (CVPR)}, July 2017, pp.
  5987--5995.

\bibitem{shvets_automatic_2018}
A.~A. {Shvets}, A.~{Rakhlin}, A.~A. {Kalinin}, and V.~I. {Iglovikov},
  ``Automatic instrument segmentation in robot-assisted surgery using deep
  learning,'' in \emph{2018 17th IEEE International Conference on Machine
  Learning and Applications (ICMLA)}, Dec 2018, pp. 624--628.

\bibitem{hu2017squeezeandexcitation}
J.~Hu, L.~Shen, S.~Albanie, G.~Sun, and E.~Wu, ``Squeeze-and-excitation
  networks,'' 2017.

\bibitem{imagenet_deng_2009}
J.~Deng, W.~Dong, R.~Socher, L.-J. Li, K.~Li, and L.~Fei-Fei, ``{ImageNet: A
  Large-Scale Hierarchical Image Database},'' in \emph{2009 IEEE Conference on
  Computer Vision and Pattern Recognition (CVPR)}, 2009.

\bibitem{ronneberger_unet_2015}
O.~Ronneberger, P.~Fischer, and T.~Brox, ``U-net: Convolutional networks for
  biomedical image segmentation,'' 2015.

\bibitem{simonyan2014deep}
K.~Simonyan and A.~Zisserman, ``Very deep convolutional networks for
  large-scale image recognition,'' 2014.

\bibitem{chollet2016xception}
F.~Chollet, ``Xception: Deep learning with depthwise separable convolutions,''
  2016.

\bibitem{mohammed2019streoscennet}
A.~Mohammed, S.~Yildirim, I.~Farup, M.~Pedersen, and {\O}.~Hovde,
  ``Streoscennet: surgical stereo robotic scene segmentation,'' in
  \emph{Medical Imaging 2019: Image-Guided Procedures, Robotic Interventions,
  and Modeling}, vol. 10951.\hskip 1em plus 0.5em minus 0.4em\relax
  International Society for Optics and Photonics, 2019, p. 109510P.

\bibitem{he2015deep}
K.~He, X.~Zhang, S.~Ren, and J.~Sun, ``Deep residual learning for image
  recognition,'' 2015.

\bibitem{peng2017large}
C.~Peng, X.~Zhang, G.~Yu, G.~Luo, and J.~Sun, ``Large kernel matters -- improve
  semantic segmentation by global convolutional network,'' 2017.

\bibitem{zhao2016pyramid}
H.~Zhao, J.~Shi, X.~Qi, X.~Wang, and J.~Jia, ``Pyramid scene parsing network,''
  2016.

\bibitem{bul_inplace_2017}
S.~R. Bulò, L.~Porzi, and P.~Kontschieder, ``In-place activated batchnorm for
  memory-optimized training of dnns,'' 2017.

\bibitem{chen_encoder_2018}
\BIBentryALTinterwordspacing
L.~Chen, Y.~Zhu, G.~Papandreou, F.~Schroff, and H.~Adam, ``Encoder-decoder with
  atrous separable convolution for semantic image segmentation,'' \emph{CoRR},
  vol. abs/1802.02611, 2018. [Online]. Available:
  \url{http://arxiv.org/abs/1802.02611}
\BIBentrySTDinterwordspacing

\bibitem{imagetoimage_isola_2016}
P.~Isola, J.-Y. Zhu, T.~Zhou, and A.~A. Efros, ``Image-to-image translation
  with conditional adversarial networks,'' 2016.

\bibitem{lin_coco_2015}
T.-Y. Lin, M.~Maire, S.~Belongie, L.~Bourdev, R.~Girshick, J.~Hays, P.~Perona,
  D.~Ramanan, C.~L. Zitnick, and P.~Doll\'{a}r, ``Microsoft coco: Common
  objects in context,'' 2015.

\end{thebibliography}
